%% file: main.tex
\documentclass[runningheads]{llncs}
\pdfoutput=1 
\usepackage[utf8]{inputenc} 
\usepackage[T1]{fontenc}    
\usepackage{hyperref}       
\usepackage{url}            
\usepackage{booktabs}       
\usepackage{amsfonts}       
\usepackage{nicefrac}       
\usepackage{microtype}      
\usepackage{xcolor}         
\usepackage{textcomp}

\usepackage{booktabs} 
\usepackage{amsmath,amssymb} 
\usepackage{graphicx}
\usepackage{multirow}
\usepackage{soul,color}
\usepackage{bm}
\usepackage{dsfont}
\usepackage{enumitem}
\usepackage{algorithmic}
\usepackage[linesnumbered,ruled,vlined]{algorithm2e}
\usepackage{booktabs}
\usepackage{url}
\usepackage{cite}
\usepackage{wrapfig, blindtext}
\usepackage{etoolbox}
\usepackage{bbm}
\usepackage{float} 
\usepackage{makecell}
\usepackage{setspace}
\usepackage{amsfonts}
\usepackage{subfig}

\newcommand{\fdl}{federated learning\ }

\DeclareMathOperator*{\argmax}{arg\,max}

\newcommand{\LLMs}{Large Language Models\ }
\newcommand{\synm}{synonymous\ }

\begin{document}
\title{Federated Prompting and Chain-of-Thought Reasoning for Improving LLMs Answering}
\titlerunning{Federated Self-consistency and CoT prompt}
%

\author{Xiangyang Liu \and Tianqi Pang  \and
Chenyou Fan}
\authorrunning{X. Liu et al.}
%

\institute{South China Normal University, Guangdong, China \\
\email{\{2022024952,2022024954\}@m.scnu.edu.cn, fanchenyou@scnu.edu.cn}}

\maketitle              
\begin{abstract}
We investigate how to enhance answer precision in frequently asked questions posed by distributed users using cloud-based Large Language Models (LLMs). Our study focuses on a typical situations where users ask similar queries that involve identical mathematical reasoning steps and problem-solving procedures. Due to the unsatisfactory accuracy of LLMs' zero-shot prompting with standalone questions, we propose to improve the distributed \synm questions using Self-Consistency (SC) and Chain-of-Thought (CoT) techniques. Specifically, we first retrieve \synm questions from a crowd-sourced database and create a federated question pool. We call these federated synonymous questions with the same or different parameters SP-questions or DP-questions, respectively. We refer to our methods as Fed-SP-SC and Fed-DP-CoT, which can generate significantly more accurate answers for all user queries without requiring sophisticated model-tuning. Through extensive experiments, we demonstrate that our proposed methods can significantly enhance question accuracy by fully exploring the synonymous nature of the questions and the consistency of the answers.


\end{abstract}

\input{sections/intro}

\input{sections/related}

\input{sections/approach}
\input{sections/experiments}

\section{Conclusion} 
We investigate the potential benefits of employing \synm queries from distributed users to enhance question-answering beyond what is achievable by a single user. Specifically, we explore the use of such queries in a federated manner by extracting two common user scenarios whereby the cloud database retrieves either SP- or DP-questions. To address these scenarios, we propose the application of self-consistency to identify the most consistent answers for SP-questions and utilize them as CoT to improve the answers provided for DP-questions. Our experimental results demonstrate that this approach yields a significant boost in performance compared to standalone zero-shot QA. 

Moving forward, future research may investigate the implementation of more realistic systems that can efficiently retrieve federated questions while also improving CoT correctness to further advance reasoning capabilities. In this study, we assumed the DP-questions have already been stored with their answers generated by LLMs, and the consistent answers have been generated. Future work can further extend to scenarios that part of the DP-questions have no answers or pseudo-answers. 




{
\bibliographystyle{splncs04}
\bibliography{egbib,fed,plm}
}

\end{document}

%% file: sections/intro.tex
\section{Introduction}

\begin{figure}[ht]
\begin{center}
\includegraphics[clip, trim=0 30 0 0, width=0.9\textwidth]{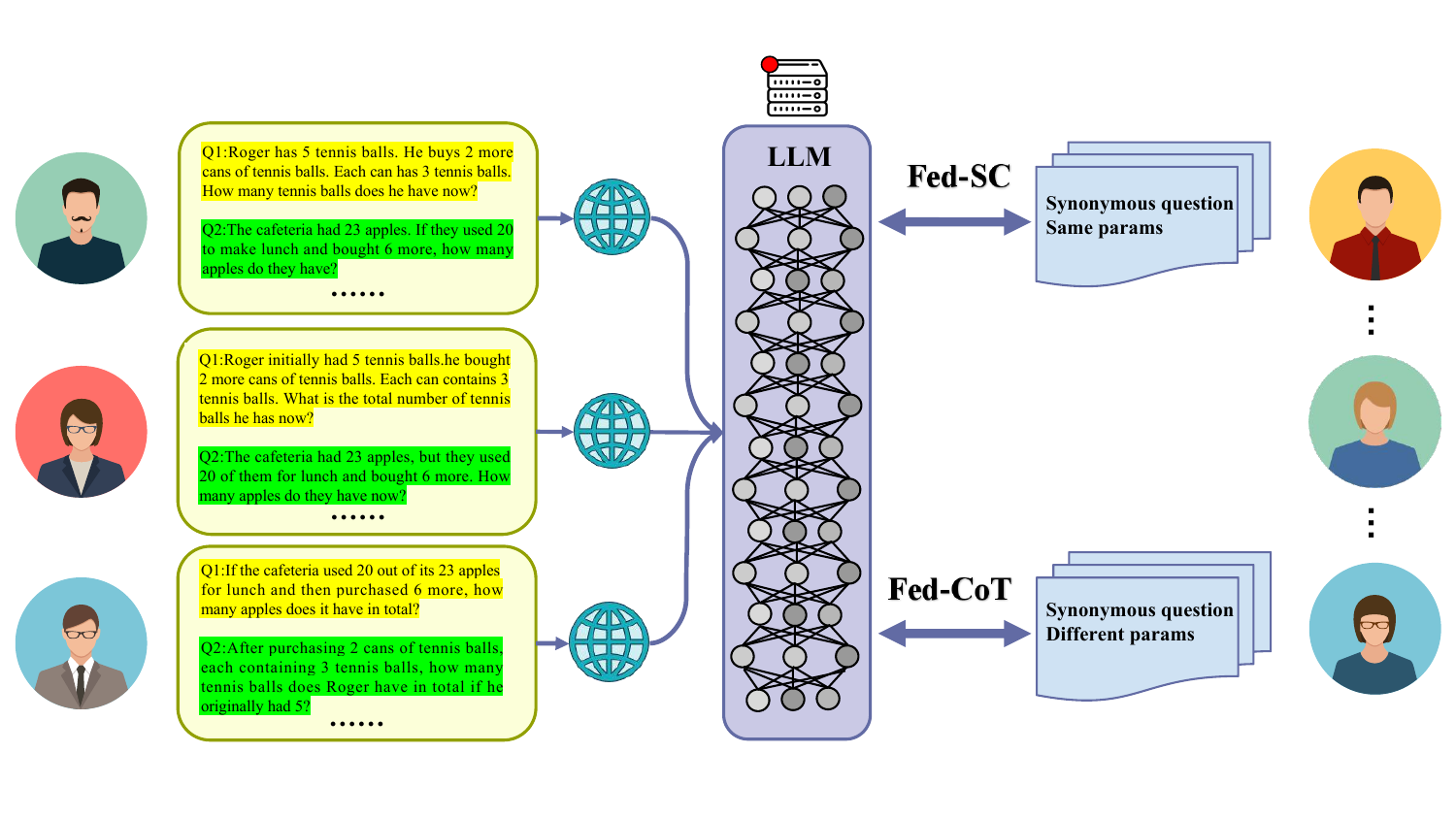}
\end{center}
\vspace{-8pt}
\caption{A general overview of our approach to dealing with federated synonymous question-answering. Our approach is categorized into two user scenarios: synonymous questions that share the same parameters, and those that have different parameters. When the parameters are the same, we utilize self-consistency to select the most commonly voted answer as the consistent response. However, for cases where the parameters are different, we amalgamate each question's consistent answer to create a Chain-of-Thought, which makes it easier for the LLM to respond to new queries. }
\label{fig:Introduction_image}
\end{figure}

Recently, \LLMs (LLMs) such as PaLM~\cite{chowdhery2022palm} and GPT family~\cite{brown2020language,ouyang2022training} 
have revolutionized the methodology of tackling natural language processing (NLP) tasks
such as sentiment analysis~\cite{wankhade2022survey}, question answering~\cite{zaib2022conversational},  text summarization~\cite{yadav2022automatic}, and reasoning on arithmetic and common sense questions~\cite{zhao2023survey}. 

\LLMs (LLMs) are highly over-parameterized with millions or billions of parameters, e.g., the GPT-3 model has about 175 Billion parameters. Due to this redundancy in model design, LLMs can represent language in a highly flexible and expressive manner by capturing the complex and structured patterns in human languages. In addition, LLMs can generate remarkably natural dialogues and accurate answers with contextual understanding, sometimes even surpassing human experts in certain tasks. For example, in arithmetic reasoning, GPT-4 achieved an accuracy rate of 92\% on the GSM8K dataset~\cite{openai2023gpt4}; in common sense reasoning, KEAR achieved an accuracy rate of 89.4\% on the CSQA dataset~\cite{xu2022human}.


We consider a practical user scenario in which a large number of users can access a cloud-deployed LLM for solving personal tasks from all places over the world. For example, more and more primary school students and their parents rely on the 
capability of LLMs for solving mathematical problems. The users often access a LLM and ask common realistic questions. For example, primary school children might ask ``Chickens and rabbits are in the same cage, a total of 35 heads, 94 feet, how many chickens and rabbits are there?'', while computer science students often ask ``How to write a QuickSort in Python?''.

Due to the complexity in task understanding and reasoning, the LLMs often return the wrong answers even given seemingly simple questions. For example, on the GSM8K dataset, the fine-tuned GPT-3 (175B) with verifier only achieves an accuracy of 55.0\%. Meanwhile, the PaLM-540B (Few-shot-CoT) only achieves an accuracy of 58.1\%.~\cite{kojima2023large} 
\emph{How to improve the question answering accuracy has become a serious challenge which decides whether LLMs can be accepted as a robust and reliable part in realistic applications.}

One commonsense is that can we crowd-source many questions and aggregate those questions to better understand some common questions. A common question might be asked frequently as its variants in the concrete parameters or rephrased formulations. For example, the Chickens-and-rabbits questions can be asked with different number of heads and feet.
Now we want to ask \emph{Can we fully utilize those similar questions to improve the question answering of the LLMs without tuning the model parameters or infringing user privacy?}

Recent progressives in \fdl (\textbf{FL})~\cite{fedavg} have proved that utilizing distributed data sources can both  preserve data privacy and enhance model training. 
In the FL paradigm, each client trains a local learning model with \emph{own data}, while a central server regularly communicates with all agents to generate a better global model through the aggregation of the local models. 

In this study, we consider improving the reasoning capacity of LLMs by better understanding crowd-sourced similar questions, from which we can explore the contextual information and improve the LLM answers substantially. 
Inspired by FL, we propose two typical scenarios when a user sends a QA request to the LLM and the LLM tries to answer with a collected question database.

\begin{itemize}[noitemsep,leftmargin=*]
\item \textbf{Synonymous Questions with Same Parameters (SP-questions)}. The cloud-deployed system retrieves from the database and finds several \synm but rephrased questions with exactly the same parameters. 
For example,\\
\emph{Q1:``If a farmer has a certain number of chickens and rabbits in a barn, and there are a total of 32 heads and 100 feet, how many chickens and how many rabbits does the farmer have?''\\
Q2:``In a barn, there are a certain number of chickens and rabbits that have a total of 32 heads and 100 feet.  how many of each animal are in the barn?''}

\item \textbf{Synonymous Questions with Different Parameters (DP-questions)}. This situation is harder as the question parameters mined in the database are different from each other. For example,\\
\emph{Q1:``If a farmer has a certain number of chickens and rabbits in a barn, and there are a total of 32 heads and 100 feet, how many chickens and how many rabbits does the farmer have?''\\
Q2:``A farmer has a total of 20 chickens and rabbits in his barn. If the total number of legs in the barn is 56, how many chickens and how many rabbits are in the barn?''}

\end{itemize}

For \textbf{SP-questions}, we propose to leverage LLMs to directly generate answers first. Then we federate the answers and apply the self-consistency~\cite{wang2023selfconsistency} technique to obtain the most voted answer for all \synm questions in the federation. We call this method \textbf{Fed-SP-SC} (Fed-SP with Self-Consistency). 

For \textbf{DP-questions}, we propose to leverage LLMs to generate consistent answers for each DP-questions first. Different from procedures of dealing SP-questions, we cannot directly agglomerate the answers since they are for different parameters. Instead, we federate them by forming the Chain-of-Thought (CoT) to provide hints to the LLMs. We append the original query to the CoT as the full prompt to obtain improved final answer. We call this technique \textbf{Fed-DP-CoT}.

Once the LLM has finished generating the answer using either Fed-SP-SC or  Fed-DP-CoT, the system will store both the questions and answers into the database. This enables the system to collect all records and leverage past records to produce re-fined answers to new queries. For questions that have been asked before with wrong answers, the system can evolve itself by correcting the  answers with self-consistency mechanism or with  more comprehensive CoT prompts.

We extensively evaluate our methods on the GSM8K and SVAMP datasets and demonstrate that the Fed-SP-SC method achieves a notable improvement in accuracy of 14-18\% over the standalone LLMs with Zero-Shot-CoT (``Let's think step by step''). Additionally, our Fed-DP-CoT method delivers an impressive increase of 10-15\% over the standalone LLMs with Zero-Shot-CoT.

We summarize our contributions in this study as follows. 
\begin{enumerate}[noitemsep,leftmargin=*]


\item We consider a practical but under-studied scenario, which is  the cloud-based LLMs are frequently asked similar and even \synm common questions from large number distributed users.

\item We abstract two main user scenarios: distributed users are querying synonymous questions that share the same parameters (SP-questions), and those that have different parameters (DP-questions).

\item We design the system to firstly federate those SP- and DP-questions first by retrieving the database. Then we propose to  utilize self-consistency methodology to select the most commonly voted answer to improve SP-question answering.  All consistent answers and CoTs will be stored back into database for further reuse.

\item We also amalgamate consistent answers to create a chain-of-thought prompt that significantly improves DP-questions answering quality.  We also design a simple disclaimer to handle noisy CoT generated from LLM answers better. 

\item Inherited from Federated Learning, our Fed-SP-SC and Fed-DP-COT methods can collaboratively enhance the question-answering process of the LLM while preserving their anonymity. There would be no data exchange or leakage among distributed users.

\end{enumerate}

%% file: sections/related.tex
\section{Related Work}

\textbf{Pre-trained Language Models (PLMs).}
Recent studies in Transformer-based language models such as ELMo~\cite{peters-etal-2018-deep} and BERT~\cite{devlin2018bert} have shown their capabilities in scaling up model sizes with pre-training methodology such as Masked Language Modeling~\cite{devlin2018bert}. 
Shortly after, several \LLMs(LLMs), e.g., the GPT family~\cite{brown2020language,ouyang2022training}, 
PaLM~\cite{chowdhery2022palm}, Jurassic-X~\cite{levine2022standing}, Megatron-Turing~\cite{shoeybi2019megatron} , LaMDA~\cite{thoppilan2022lamda},LLaMA~\cite{touvron2023llama}, have been emerging with huge amount of parameters of up to 100B-5000B parameters. They have shown great advantages in language modeling tasks, such as arithmetic reasoning, commonsense reasoning, symbolic reasoning and natural language inference.

However, PLMs are still like black boxes which lack of explanation. Some recent studies made efforts towards unveiling the power of those LLMs. The proposal of the concept of the \emph{Chain-of-Thought} (CoT)~\cite{wei2022chain} indicates that incorporating intermediate reasoning steps can lead to a significant improvement in the performance of large language models on reasoning tasks. 
The proposal of the \emph{Self-consistency}~\cite{wang2023selfconsistency} suggest that aggregating multiple reasoning paths, rather than relying on greedy decoding, can lead to further improvements in the accuracy of models on reasoning tasks.
LMSI~\cite{huang2022large} provides a demonstration of how large language models can achieve self-improvement by utilizing only unlabelled datasets.

However, \textbf{it is unknown how to apply proper pre-training to distributed learning scenarios}, due to substantial differences between centralized large model deployment and distributed query demands.
In this study, we adopt the recent popular distributed machine learning methodology called 
\emph{Federated Learning}~\cite{fedavg,zhao2018federated,fan2021fedfsl,fan2022private} (FL) to fully explore the potentiality of \LLMs to tackle frequently asked questions while preserving data privacy for the users. 
The FL provides a way of learning models over a collection of distributed devices while keeping data locality. However, classical FL studies assumed the agents in FL can own copies of local models while receiving updates from centralized model. In contrast, we focus on a practical scenario that the clients can only query answers from centralized \LLMs without owning any local model, due to the practical situations that \LLMs are simply too large and computational extensive to be deployed locally.

%% file: sections/approach.tex
\section{Scenarios and Approaches}
In this section, we describe the federated scenarios that distributed users query the LLMs with similar (but not exact the same) questions. We identify two types of questions and discuss them in details.

\subsection{Basic Concepts}

\textbf{Chain-of-Thought (CoT)}~\cite{wei2022chain} is a series of generated intermediate reasoning texts that can be added to the original prompts. CoT is proposed for enhancing the capability of language models to perform various reasoning tasks by allowing LLMs to decompose complex problems into intermediate steps that could be solved well step-by-step. Chain-of-thought prompting, i.e. prompting LLMs with CoT, is a simple and practical method for improving the reasoning tasks readily with no additional efforts of tuning the original LLMs. CoT prompting has shown improved reasoning results on arithmetic, commonsense, and symbolic reasoning tasks. 

\textbf{Self-Consistency (SC)}~\cite{wang2023selfconsistency} is a decoding strategy that enhances language model reasoning with voting ensemble. SC first samples a diverse set of answers as reasoning paths of a question, rather than only the greedy path. By exploring multiple paths, SC is capable of identifying the most consist answer as the final answer by majority voting, i.e., the most voted answer of the LLM is taken as the final answer. Compared with a single-path reasoning, SC ensembles answers to improve accuracy and filters out noises or outliers. 
SC has also been widely explored in reasoning and QA tasks~\cite{wang2023selfconsistency}.

\textbf{Majority voting(MV)}\cite{Ongaro2014Consensus} is a commonly used method in statistical decision theory that involves aggregating the opinions or decisions of multiple individuals or models, typically by selecting the option with the highest frequency of agreement among the voters.

\subsection{Synonymous Questions with Same Parameters (SP-questions)}
We consider a cloud-based LLM system which accepts queries from distributed users. The first practical user scenario that we consider is as follows. Given a user query, we can retrieve from the cloud database several \textbf{synonymous questions with same parameters (SP-questions)}. 


\begin{table}[ht]
    \centering 
    \vspace{-5pt}
    \caption{Examples of \synm SP-questions and answers.  }
    \begin{tabular*}{\hsize}{@{}@{\extracolsep{\fill}}l@{}}
         \toprule
         \emph{Example1:}\\
         \midrule
         \emph{\textbf{Q1:} ``If a farmer has a certain number of chickens and rabbits in a barn, and }\\\emph{there are a total of $32$ heads and 100 feet, how many chickens and how many}\\\emph{rabbits does the farmer have?''}\\
         \emph{\textbf{A1:} ``The farmer has $24$ chickens and 8 rabbits.''} \textcolor{red}{(wrong)}\\
         \midrule
         \emph{\textbf{Q2:} ``In a barn, there are a certain number of chickens and rabbits that have}\\\emph{a total of $32$ heads and $100$ feet.  how many of each animal are in the barn?''}\\
         \emph{\textbf{A2:} ``Let $x=$ the number of chickens and $y=$ the number of rabbits. We can}\\\emph{ set up the following system of equations:$x + y = 32  (heads)$, $2x + 4y = 100$}\\\emph{ (feet), Solving this system of equations, we get $x = 20$ and $y = 12$.Therefore,}\\\emph{ there are $20$ chickens and $12$ rabbits in the barn.''} \textcolor{red}{(wrong)}\\
         \toprule
         \emph{Example2:}\\
         \midrule
         \emph{\textbf{Q1:} ``James writes a $3$-page letter to $2$ different friends twice a week.How many}\\\emph{pages does he write a year?''}\\
         \emph{\textbf{A1:} ``James writes $3$ pages to $2$ different friends twice a week, which is $24$ pages} \\\emph{ a month and $288$ pages a year.''} \textcolor{red}{(wrong)}\\
         \midrule
         \emph{\textbf{Q2:} ``If James writes a $3$-page letter to two different friends twice per week,what} \\\emph{ is the total number of pages he produces every year?''}\\
         \emph{\textbf{A2:} ``James writes two $3$-page letters twice per week. There are $52$ weeks in a }\\\emph{year. Therefore, James produces a total of $312$ pages every year($2 * 3 * 52=312$).''} \\ \textcolor{red}{(wrong)}\\
        \bottomrule
    \end{tabular*}
    \label{tab:SP-question-example}
\end{table}

For complex reasoning tasks, the LLMs may provide unreliable answers to the questions. We provide two failure cases in Table.~\ref{tab:SP-question-example}. We found that in both examples, Q1 and Q2 are \synm while each of them gets a wrong answer from LLM. We summarize the difficulties of tackling the SP-questions as follows: 
\begin{enumerate}[noitemsep,leftmargin=*]
\item Most LLMs have unsatisfying accuracy in solving reasoning problems in zero-shot way, i.e.,  prompting the LLMs with questions directly without giving other information.
\item LLMs tend to under-perform when understanding complex problems involving multiple reasoning steps, such as the arithmetic problems given above.
\end{enumerate}

Thus, our task is to \emph{fully explore the SP-questions as a federation which can enhance the answer quality together}, instead of dealing them separately. To this end, we propose a technique named \textbf{Fed-SP-SC} (Federated SP-questions with Self-Consistency) for answering the questions with the self-consistency technique mentioned above. Fed-SP-SC can improve the zero-shot accuracy of the answers by eliciting answers from multiple \synm questions and make a majority vote to disclose the most likely answer. 

Concretely, we query the database using the user's prompt to match SP-questions in the database. Note that here we assume we can retrieve the SP-questions which are just rephrased with \synm and same parameters.

Next, we generate the answers with LLMs by zero-shot prompting. For SP-questions, these answers are presumably same. Assuming that we have generated a total of \emph{n} 
answers of \synm questions during the Fed-SP-SC process, we can ensure the consistency with SC procedure, i.e., we make a majority vote and select the most voted answer $A^{SC}$ as the final answer of all SP-questions, as below.
\begin{equation}
    A^{SC} \leftarrow \argmax_{A \in \mathcal{A}} \sum \mathbf{1}[A==A_i], \ \ 
\forall i=1,...,n  \ .
\end{equation}

Intuitively, the majority voting filters out outliers and noisy rephrased questions. In addition,
the most voted answer is the agreement of multiple reasoning paths from multiple rephrased SP-questions, thus is more likely to be better than a single prompted answer decoded from a single reasoning path.  

In our experiments, we demonstrate that Fed-SP-SC achieves a 17.5\% improvement in accuracy on the GSM8K dataset and a 14\% improvement on the SVAMP dataset in Table \ref{tab:Fed_SP_SC_result}. In a practical system, we can further store these user prompts and the SC-selected answer back into the database.



\begin{figure}[ht]
\begin{center}
\includegraphics[clip, trim=0 20 0 0, width=0.9\textwidth]{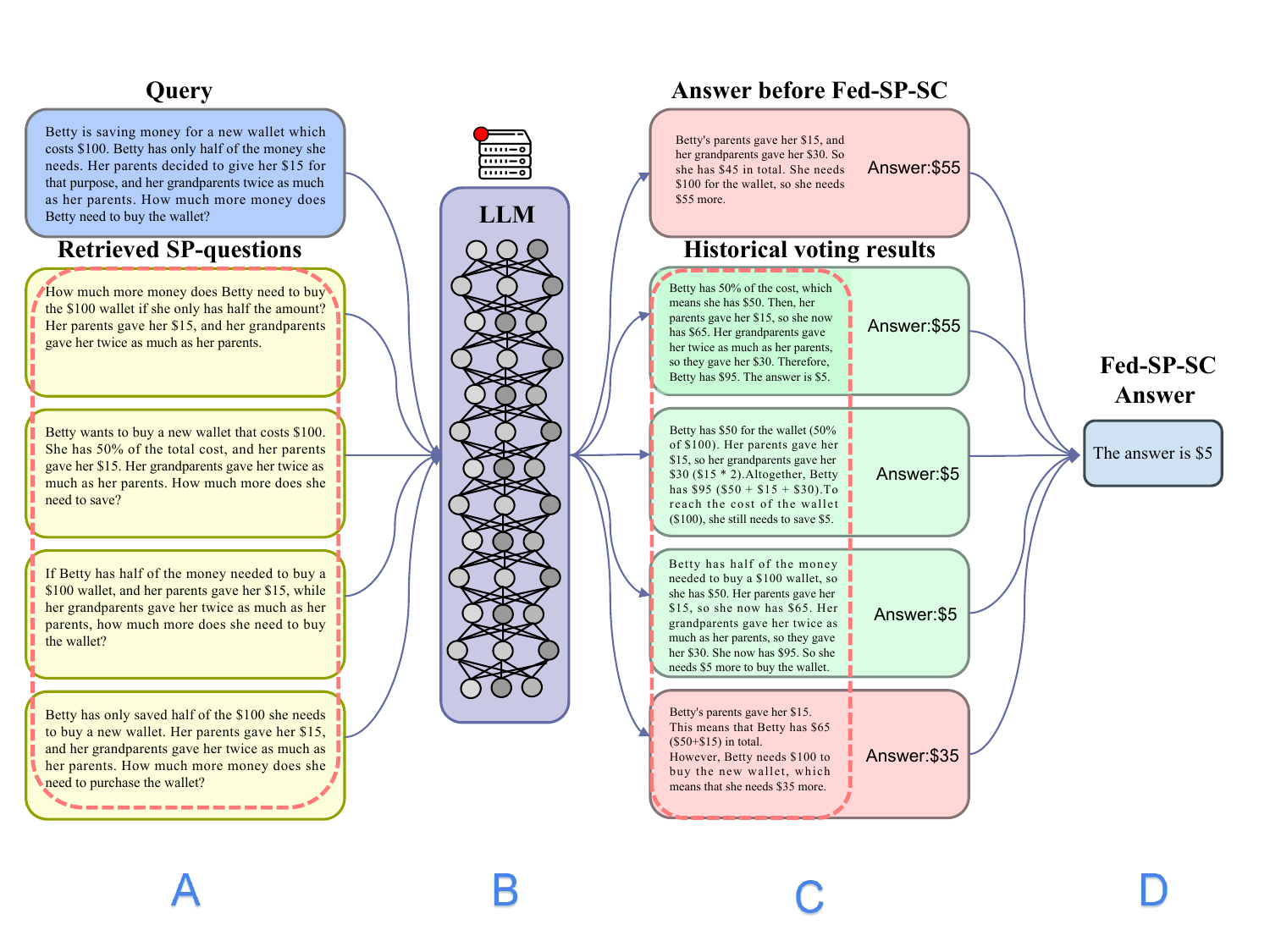}
\end{center}
\vspace{-8pt}
\caption{The illustration of performing Fed-SP-SC for answering \synm SP-questions. $(A \rightarrow B)$: When receiving the user's query, the LLM retrieves \synm SP-questions from the centralized database. $(B \rightarrow C)$: The LLM generates the answers with zero-shot prompting for the query and combines the retrieved SP-questions' answers from the database for a majority vote to ensure self-consistency. $(C \rightarrow D)$: The most voted answer is returned to the user as the best answer. The database could store the query and answer pair back to the database, caching for later retrieval. This procedure can grow the database quickly by gathering distributed user queries. }
\label{fig:Fed-SC}
\vspace{-6pt}
\end{figure}


 \subsection{Synonymous questions with Different Parameters (DP-questions)}
We now describe the second scenario which is named \textbf{\synm questions with different parameters (DP-questions)}, which is broader and more practical. 
Based on the user query question, the cloud-deployed system searches and retrieves from the database for questions with same meanings but may have different parameters.  

DP-questions are more practical yet harder than SP-questions as the question parameters retrieved from the database are different. We show two exemplary questions Q1 and Q2 below which have the same meaning yet with different parameters \emph{heads} and \emph{feet} in Table \ref{tab:DP-question-example}. 
\begin{table}[!ht]
    \centering 
    \caption{Two examples of DP-questions. \emph{Q1} and \emph{Q2} are synonymous but with different question parameters.
}
    \begin{tabular*}{\hsize}{@{}@{\extracolsep{\fill}}l@{}}
         \toprule
         The specific description of \emph{Q1} and \emph{Q2:}\\
         \midrule
         \emph{\textbf{Q1: } ``If a farmer has a certain number of chickens and rabbits in a barn and},\\\emph{there are a total of 32 heads and 100 feet, how many chickens and how many}\\\emph{rabbits does the farmer have?''}\\
         \emph{\textbf{Q2: } ``A farmer has a total of 20 chickens and rabbits in his barn. If the total}\\\emph{number of legs in the barn is 56, how many chickens and how many rabbits} \\\emph{are in the barn?''}\\
        \bottomrule
    \end{tabular*}
    \label{tab:DP-question-example}
\end{table}

Note that tackling DP-questions would face all the difficulties of SP-questions, and would have additional obstacles as summarized below: 
\begin{enumerate}
\item There is no uniform ground-truth for DP-questions in the database, since each one has different parameters. 
\item Similarly, we cannot apply self-consistency (SC) directly to improve accuracy due to different parameters.
\item If we apply Chain-of-Thought (CoT) together to the original questions as enhanced prompts, we cannot guarantee the correctness of the CoT. Incorrect CoT may even harm the answering accuracy.
\end{enumerate}

To tackle the above challenges, we propose the \textbf{Federated questions of Different Parameters with Chain-of-Thought} (Fed-DP-CoT) technique to leverage existing answers of DP-questions in CoT forms to improve new query answering. 
\begin{figure}[!ht]
\begin{center}
\includegraphics[clip, trim=0 60 0 0, width=0.9\textwidth]{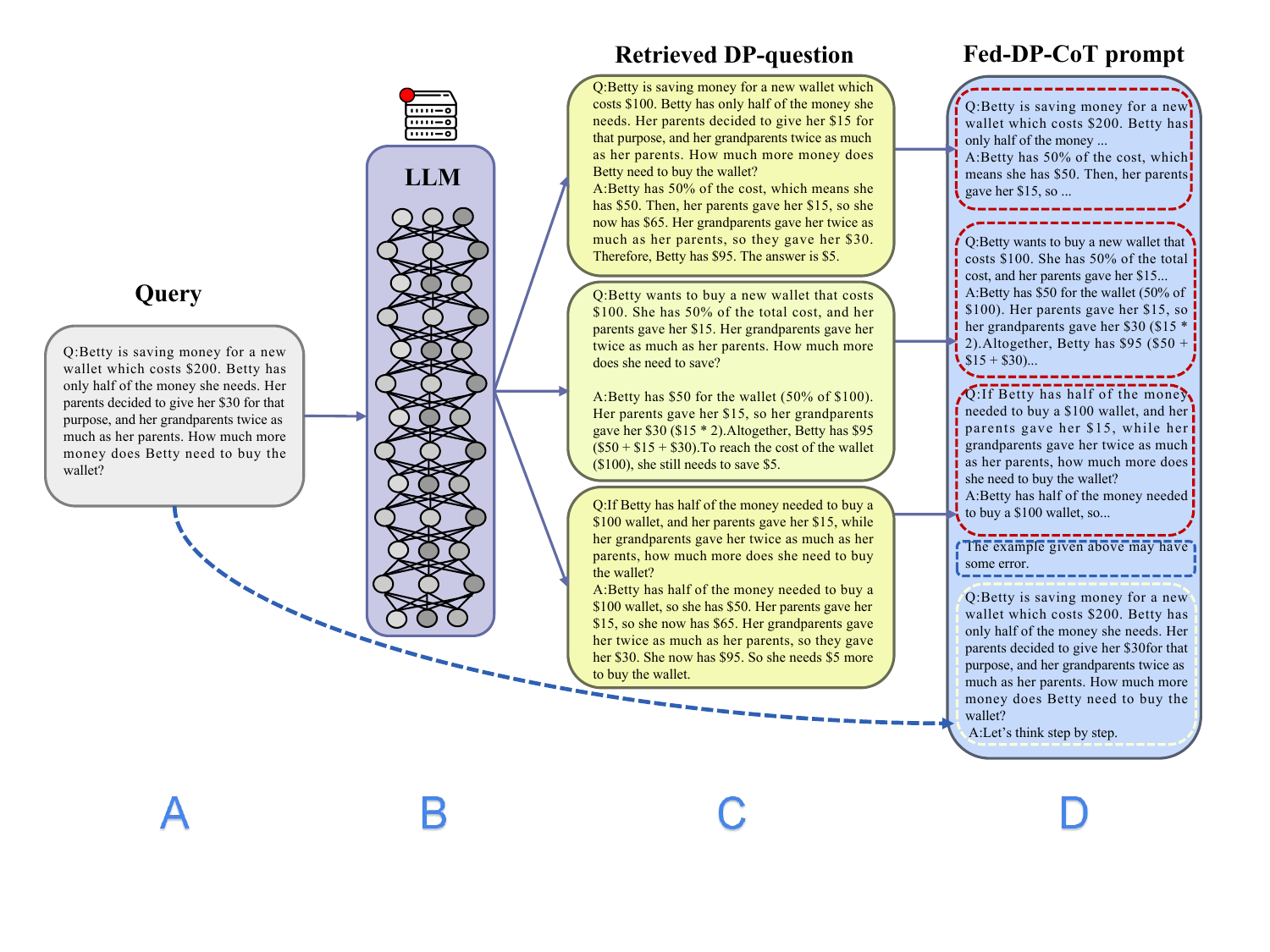}
\end{center}
\vspace{-5pt}
\caption{The illustration of performing Fed-DP-CoT. 
$(A \rightarrow B)$: DP-questions are firstly retrieved from the centralized user query  database. $(B \rightarrow C)$: The system selects SP-questions with consistent answers after applying Fed-SP-SC as DP-questions and retrieves consistent answers as ``pseudo-labels''. $(C \rightarrow D)$: The system concatenates questions and pseudo-labels, adds a pseudo-label disclaimer such as ``The examples may have errors.'' after, and finally appends the original user query and Zero-Shot-CoT to form the complete CoT prompt. }
\label{fig:Fed-CoT}
\vspace{-6pt}
\end{figure}


When a user starts querying the cloud-based LLM service, the cloud system performs query-question retrieval first. The system matches several questions with the highest similarity in the database. Generally, these  the retrieved questions are of different parameters (DP-questions). We design the system  to perform Fed-DP-CoT for understanding DP-questions.

We consider a practical case that these DP-questions have \emph{pseudo-labels} generated by self-consistency majority voting in the Fed-SP-SC processes.
We call these labels ``pseudo-labels'' as they are not actual ground-truth labels.


Then we utilize these DP-questions with pseudo-labels together as CoT for the original query-question. 
To be specific, we concatenate DP-questions with their answers as a single prompt, followed by the error disclaimer
``The examples given above may contain errors , please think more carefully. '' at the end of this prompt  as the complete prompt.
The error disclaimer reminds the LLMs that the answers in CoT are pseudo-labels and could be incorrect. We found this simple practice can boost performance by approximately 2\%.
Finally, we use the entire disclaimed CoT as a prefix to the user's query prompt for the LLMs to provide the final answer.


%% file: sections/experiments.tex
\section{Experiment}
\label{sec:exp}

We evaluate our proposed \textbf{Fed-SP-SC} and \textbf{Fed-DP-CoT} methods on benchmark datasets with simulated user scenarios such that SP- and DP-questions are retrieved to improve over standalone question answering.

We compare our methods with \emph{Zero-Shot-CoT}~\cite{brown2020language}, which refers to adding \emph{``Let's think step by step.''} to prompt as a composite prompt, such as ``[Question] A:Let's think step by step.'' 

\subsection{Datasets}
\label{sec:dataset}

\textbf{Grade School Math (GSM8K)} is a math dataset with 7,473 training and 1,319 testing examples of grade-school-level word problems\cite{Cobbe2021TrainingVT}. These math problems typically require two to eight calculation steps to arrive at a final answer, as shown in Fig.\ref{fig:Fed-SC}. GSM8K is widely used as a benchmark dataset for testing the arithmetic and commonsense reasoning capacities of LLMs~\cite{huang2022large, kojima2023large}.


\textbf{Simple Variations on Arithmetic Math word Problems (SVAMP)} is a dataset of simple arithmetic math word problems~\cite{patel2021nlp} with around 6,000 samples. Each data instance has a short story and a question about unknown quantities. SVAMP provides a benchmark test set for comparing the textual understanding and reasoning abilities of LLMs, which is widely compared in recent studies~\cite{patel2021nlp, wei2022chain, wang2023selfconsistency}.


In practice, we utilized the OpenAI's API~\footnote{https://platform.openai.com/docs/models} text-davinci-002 and text-davinci-003. We selected text-davinci-003 for the GSM8K dataset as text-davinci-002 performed very poorly. Similarly, we used text-davinci-002 for the SVAMP dataset as text-davinci-003 had an overly high accuracy rate on this dataset.


\subsection{Results of Fed-SP-SC}
\label{sec:fed-sp-sc_result}

\begin{figure}[ht]
\centering
\includegraphics[width=1\textwidth]{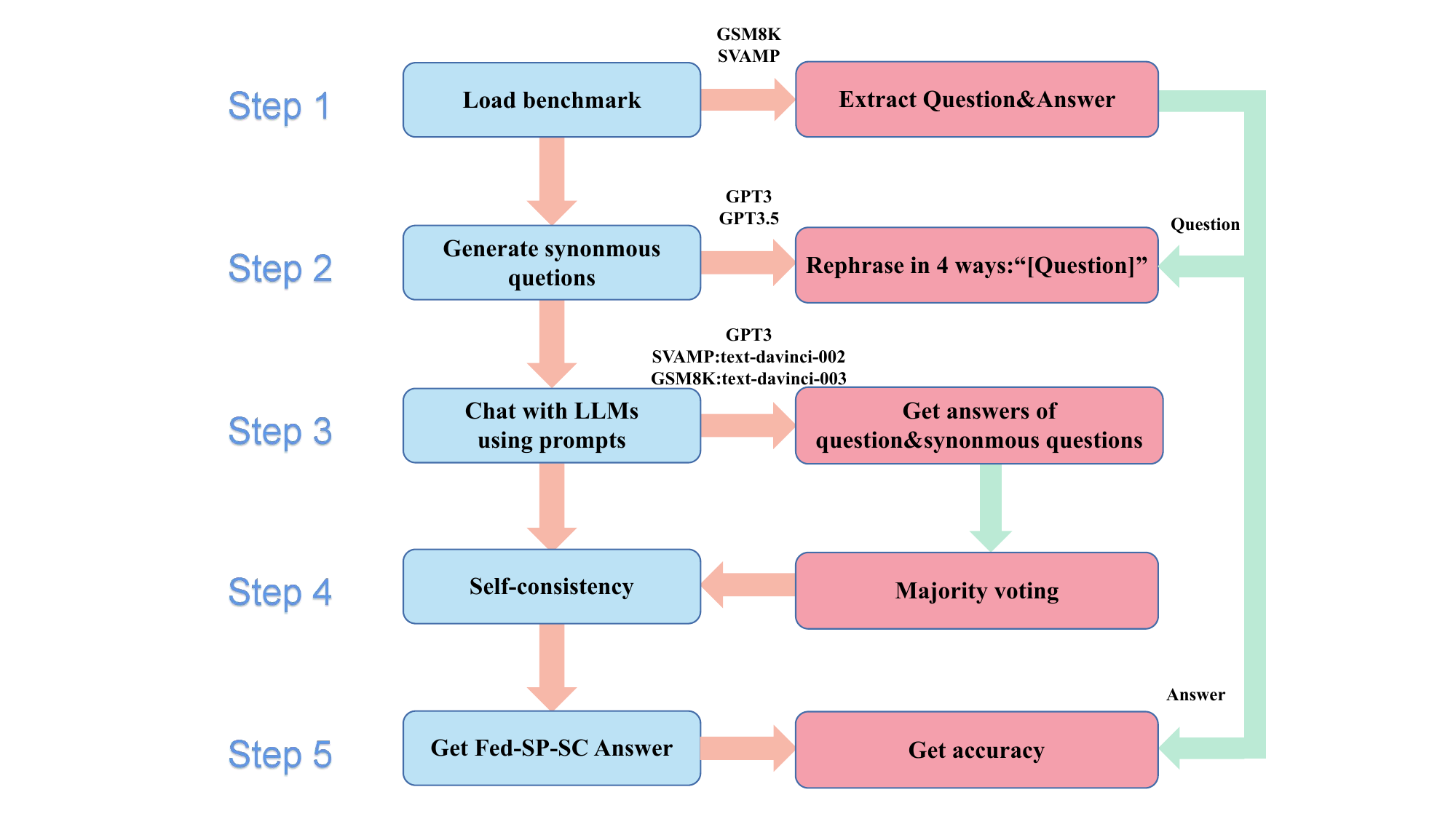}
\caption{The experiment of Fed-SP-SC contains five steps: (1) Load the GSM8K and SVAMP datasets as our benchmark and extract the questions and answers in the dataset; (2) Add each original question a same prompt prefix ``Rephrase in 4 ways: [QUESTION]'' to generate SP-questions; (3) Prompt both the original and rewritten questions to the LLMs to obtain their respective answers; (4) Use the majority vote for self-consistency;  (5) Get the answer generated by Fed-SP-SC and compare it with the answer in the dataset to determine the accuracy rate.}
\label{fig:image}
\end{figure}

As a kind reminder, in the following discussions, \textbf{SP-questions} stand for \emph{a set of rephrased \synm questions with same parameters.} Differently, \textbf{DP-questions} stand for \emph{a set of rephrased \synm questions with different parameters. }
We now describe the experiment procedures shown in Fig.~\ref{fig:image}.

[\textbf{SP-questions generation}]. We first generate four SP-questions for each of the original question with both OpenAI GPT-3~\cite{brown2020language} and GPT-3.5~\cite{ouyang2022training}, respectively. Concretely, we add each original question a same prompt prefix \textit{``Rephrase in 4 ways: [ORIGINAL QUESTION]''}, then we collect the generated answers.

[\textbf{SP-questions answering}].
We query the cloud-deployed LLM for answering rephrased questions generated as above.
Specifically, we obtain the improved Zero-Shot-CoT answering with the magic sentence \textit{``Let's think step by step''} as the prompt. 


We first examine the performance of our proposed Fed-SP-SC which deals with SP-questions with the Self-consistency technique. 
We conducted experiments on GSM8K and SVAMP and report the results below.

\begin{table}[ht]
\caption{Fed-SP-SC results}
\label{tab:Fed_SP_SC_result}
\centering
\begin{tabular}{l|c | c | c}
\toprule
Data\textbackslash{}Method & Zero-Shot-CoT & \begin{tabular}[c]{@{}c@{}}Fed-SP-SC \\ (GPT-3 Gen.)\end{tabular} & \begin{tabular}[c]{@{}c@{}}Fed-SP-SC \\ (GPT-3.5 Gen.)\end{tabular} \\ 
\toprule
\multicolumn{1}{c|}{GSM8K} & \multicolumn{1}{c|}{52.5\%} & \multicolumn{1}{c|}{62.7\%}  & \multicolumn{1}{c}{70.6\%}    \\
\multicolumn{1}{c|}{SVAMP} & \multicolumn{1}{c|}{77.2\%} & \multicolumn{1}{c|}{86.3\%}  & \multicolumn{1}{c}{91.1\%}  
\\ 
\toprule
\end{tabular}
\end{table}

We show accuracy of self-consistency after obtaining results from different phrasings of the \synm question on GSM8K and SVAMP in Table~\ref{tab:Fed_SP_SC_result}. 
We have the following observations.

\begin{itemize}[noitemsep,leftmargin=*]
\item \emph{Fed-SP-SC can improve answering accuracy} of LLMs by federating multiple SP-questions through self-consistency. 
\item \emph{Fed-SP-SC(GPT-3.5 Gen.) performs best on the GSM8K and SVAMP datasets,} improved the performance by $17.5\%$ and $14\%$ on the GSM8K and SVAMP datasets, respectively.
\item \emph{The quality of the \synm questions can affect the accuracy significantly}, as seen in the larger improvement from the \synm questions generated by GPT-3.5 compared to GPT-3.
\end{itemize}

\subsection{Results of Fed-DP-CoT}
\label{sec:fed_cot_result}

\begin{figure}[ht]
\centering
\includegraphics[width=1\textwidth]{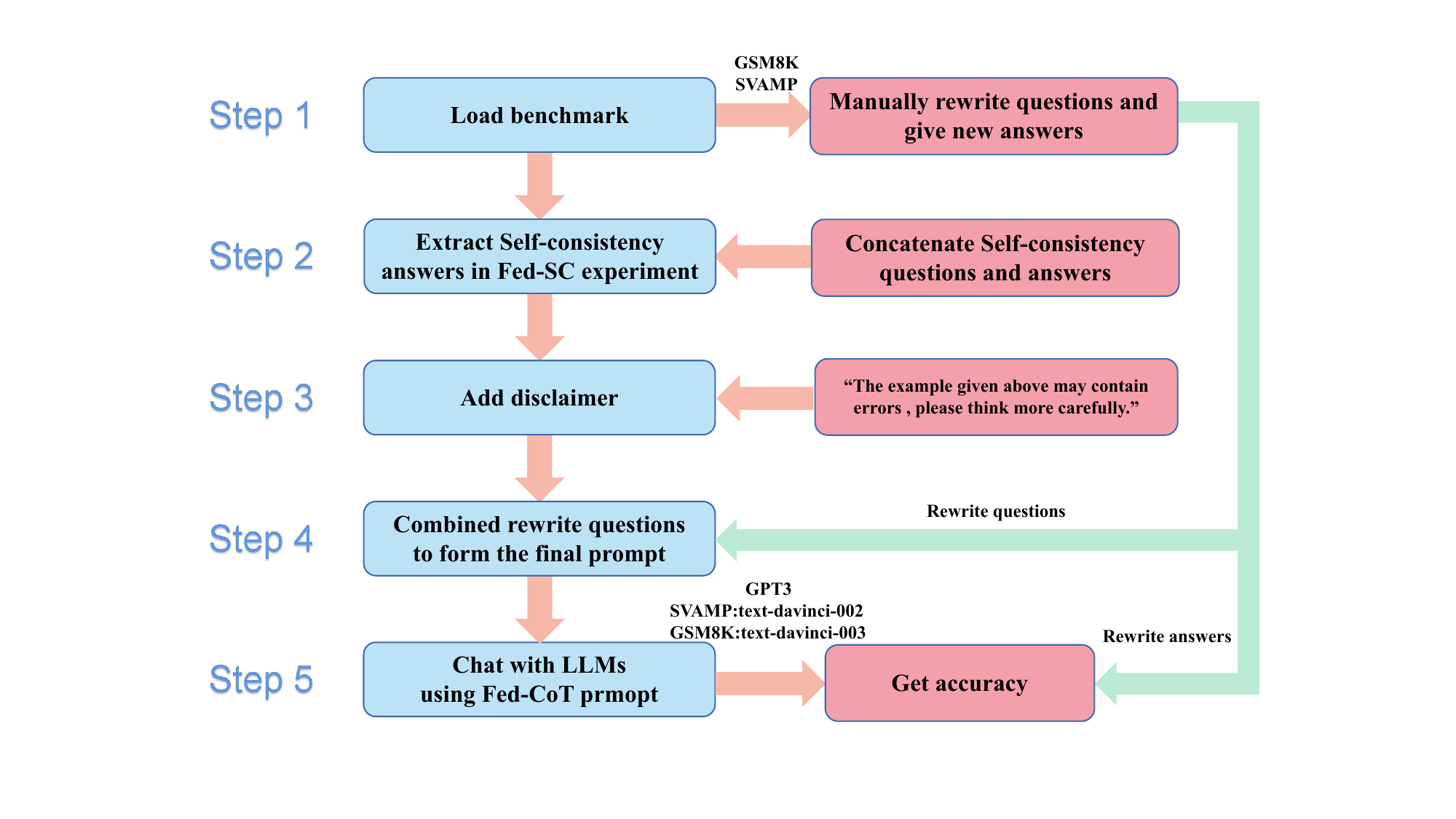}
\caption{The experiment of Fed-DP-CoT contains five steps:(1) Form a new test set by rephrasing the questions using different parameters in the benchmark manually and providing answers; (2) Extract consistent questions and answers in Fed-SP-SC experiment; (3) Add a disclaimer to form the CoT prompt; (4) Add the rephrased questions after the CoT prompt; (5) Prompt LLMs with entire CoT prompt and compared the answers with the rephrased answers for evaluation. }
\label{fig:Fed-DP-CoT-flow-chart}
\end{figure}




\begin{table}[htp]
\caption{Fed-DP-CoT results.}
\label{tab:Fed_CoT_result}
\centering
\begin{tabular}{l|c| c| c}
\toprule
Setting\textbackslash{}Method & Zero-Shot-CoT & \begin{tabular}[c]{@{}c@{}}Fed-DP-CoT\\ (GPT-3 Gen.)\end{tabular} & \begin{tabular}[c]{@{}c@{}}Fed-DP-CoT\\ (GPT-3.5 Gen.)\end{tabular} \\ 
\toprule
\multicolumn{1}{c|}{GSM8K} & \multicolumn{1}{c|}{48.3\%} & \multicolumn{1}{c|}{59.2\%}  & \multicolumn{1}{c}{62.5\%}    \\
\multicolumn{1}{c|}{SVAMP} & \multicolumn{1}{c|}{76.5\%} & \multicolumn{1}{c|}{82.4\%}  & \multicolumn{1}{c}{85.7\%}  
\\ 
\toprule
\end{tabular}
\end{table}

We report results of Fed-DP-CoT on GSM8K and SVAMP in Table~\ref{tab:Fed_CoT_result}, and compare with the baseline Zero-Shot-CoT.

\begin{itemize}[noitemsep,leftmargin=*]
\item \emph{Fed-DP-CoT can improve the performance.} Compared to Zero-Shot-CoT, CoT Prompt(GPT-3 Gen.) and CoT Prompt(GPT-3.5 Gen.) improve by approximately 10.9\%-14.2\% and 6.6\%-10\% respectively on the datasets GSM8K and SVAMP.
\item \emph{Fed-SP-SC performs better than Fed-DP-CoT.}  The results of Fed-SP-SC (GPT-3 Gen.) and Fed-SP-SC (GPT-3.5 Gen.) on the GSM8K and SVAMP datasets are both higher than Fed-DP-CoT (GPT-3 Gen.) and CoT Prompt (GPT-3.5 Gen.), with an approximate improvement of 5\%.
\item \emph{Less significant performance difference between GPT-3 Gen. and GPT-3.5 Gen. compared to Fed-SP-SC experiment.} The reason for this is the disparity in parameters employed, coupled with the lack of emphasis on synonym usage in the CoT prompt.
\end{itemize}

\subsection{Ablation studies}
\label{sec:ablation}

\begin{figure}[ht]
\centering
\subfloat[GSM8K]{
\includegraphics[width=0.48\textwidth]{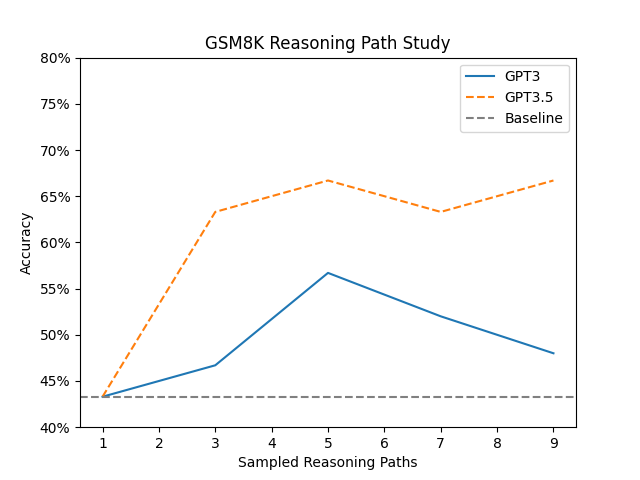}
\label{fig:GSM8K-ablation-study}
}
\hfill
\subfloat[SVAMP]{
\includegraphics[width=0.48\textwidth]{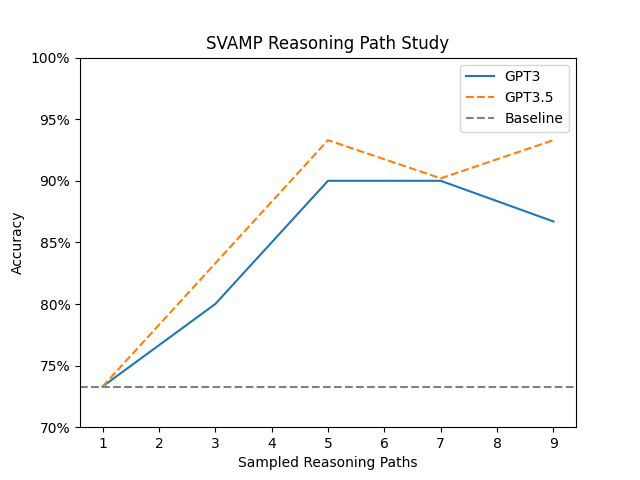}
\label{fig:SVAMP-ablation-study}
}
\caption{Ablation study of choice of sampled reasoning paths.}
\label{fig:ablation-study}
\end{figure}

\textbf{The number of reasoning paths for self-consistency.}
We study the effect of using different number of sampled reasoning paths for Fed-SP-SC (Sec.~\ref{sec:fed-sp-sc_result}) to apply self-consistency. We conduct hyper-parameter search with a subset of the data  for this ablation study due to the limits of accesses of the OpenAI API.  

We vary the number of sampled reasoning paths of \synm questions from one to nine. Figure \ref{fig:ablation-study} shows that increasing the number of sampled reasoning paths of the \synm questions does not always improve the accuracy of the model. 

In the line chart, as the number of sampled reasoning paths increases from one to five, the accuracy rate gradually increases.
However, when the number of \synm questions exceeds five, the accuracy of the model starts to decrease. 

We speculate that this is because introducing \synm questions also introduces noisy phrases, causing a deviation in the semantic meaning of the original questions. This deviation is particularly evident in synonymous questions generated by GPT-3 (blue lines), while the generation results of GPT-3.5 (orange lines) exhibit stronger robustness.

\begin{table}[ht]
\caption{GSM8K disclaimer ablation. }
\label{tab:disclaimer ablation}
\centering
\begin{tabular}{c|c|c|c}
\toprule
Method\textbackslash{}Setting & \begin{tabular}[c]{@{}c@{}}Zero-shot\\-CoT\end{tabular}        & \begin{tabular}[c]{@{}c@{}}Fed-DP-CoT\\ (GPT-3 Gen.) \end{tabular} & \begin{tabular}[c]{@{}c@{}}Fed-DP-CoT\\ (GPT-3.5 Gen.) \end{tabular} \\
\toprule
w/o disclaimer           & 48.3\%           & 57.7\%            & 60\% \\                                   
w/ disclaimer           & NA                 & 59.2\%            & 62.5\%  \\ \toprule                           
\end{tabular}
\end{table}

\textbf{Disclaimer}  We investigate whether the disclaimer is effective of correcting noisy CoTs in this ablation experiment. As Zero-Shot-CoT does not employ pseudo-labels, we do not conduct disclaimer ablation on it. Table \ref{tab:disclaimer ablation} compares the DP-questions answering accuracy with disclaimer or without disclaimer. We observe that the addition of a disclaimer in the questions and answers generated by GPT-3 resulted in an increase in accuracy from 57.7\% to 59.2\% for the Fed-DP-CoT task. Similarly, in the case of questions and answers generated by GPT-3.5, the accuracy increase from 60\% to 62.5\%. These results indicate that the use of a simple disclaimer can potentially improve the accuracy of LLMs by approximately 2\% for the Fed-DP-CoT task. We postulate that the improvement in accuracy may be attributed to the fact that the disclaimer prompts LLMs to be careful of the pseudo-labels and self-examine the reasoning steps.